\RequirePackage{cmap}

\documentclass[conference]{IEEEtran}[2015/08/26]

\usepackage[T1]{fontenc}
\usepackage[utf8]{inputenc}

\usepackage[english]{babel}

\usepackage{mwe}

\usepackage[zerostyle=b,scaled=.75]{newtxtt}

\usepackage{xtab,booktabs}

\usepackage{graphicx}

\usepackage{listings}
\usepackage{upquote}

\usepackage{balance}

\usepackage[square,comma,numbers,sort&compress]{natbib}

\usepackage{etoolbox}
\makeatletter
\patchcmd{\NAT@test}{\else \NAT@nm}{\else \NAT@hyper@{\NAT@nm}}{}{}
\makeatother

\usepackage[hyphens]{url}

\usepackage{hyperref}

\hypersetup{hidelinks,
  colorlinks=true,
  allcolors=black,
  pdfstartview=Fit,
  breaklinks=true
}

\usepackage[all]{hypcap}

\usepackage[caption=false,font=footnotesize]{subfig}

\usepackage{stfloats}

\usepackage{longtable}
\usepackage[table,xcdraw]{xcolor}

\begin{document}

\title{Autonomous Grasping On Quadruped Robot With Task Level Interaction}






\author{%
Muhtadin$^{1}$$^{,}$$^{2}$,
Mochammad Hilmi Rusydiansyah$^{2}$,
Mauridhi Hery Purnomo$^{1}$$^{,}$$^{2}$,
I Ketut Eddy Purnama$^{1}$, \\
Chastine Fatichah$^{3}$ \\
\small $^{1}$Department of Electrical Engineering, Institut Teknologi Sepuluh Nopember, Surabaya, Indonesia \\
\small $^{2}$Department of Computer Engineering, Institut Teknologi Sepuluh Nopember, Surabaya, Indonesia \\
\small $^{3}$Department of Informatics, Institut Teknologi Sepuluh Nopember, Surabaya, Indonesia \\
Email: muhtadin@its.ac.id,  rusydiansyahhilmi@gmail.com, hery@ee.its.ac.id, ketut@te.its.ac.id, chastine@if.its.ac.id
}\maketitle
\renewcommand\abstractname{Abstract}

\begin{abstract}

Quadruped robots are increasingly used in various applications due to their high mobility and ability to operate in diverse terrains. However, most available quadruped robots are primarily focused on mobility without object manipulation capabilities. Equipping a quadruped robot with a robotic arm and gripper introduces a challenge in manual control, especially in remote scenarios that require complex commands. This research aims to develop an autonomous grasping system on a quadruped robot using a task-level interaction approach. The system includes hardware integration of a robotic arm and gripper onto the quadruped robot’s body, a layered control system designed using ROS, and a web-based interface for human-robot interaction. The robot is capable of autonomously performing tasks such as navigation, object detection, and grasping using GraspNet. Testing was conducted through real-world scenarios to evaluate navigation, object selection and grasping, and user experience. The results show that the robot can perform tasks accurately and consistently, achieving a grasping success rate of 75\% from 12 trials. Therefore, the system demonstrates significant potential in enhancing the capabilities of quadruped robots as service robots in real-world environments.

\end{abstract}

\renewcommand\IEEEkeywordsname{Keywords}


\section{Introduction}
\label{sec:introduction}

Quadruped robots have garnered growing attention as versatile solutions in complex environments. These four-legged robots are designed to emulate the locomotion of animals like dogs or horses \cite{akira_fukuhara_14f4446c},\cite{xijun_he_1853cb30},\cite{marco_hutter_74e8b5a1} . With this structure, quadruped robots offer better stability than bipedal robots and greater mobility over uneven terrain compared to wheeled robots \cite{LeiWu},\cite{yixuan_liu_e938e4ee}, \cite{azhar_aulia_saputra_3c28949b}. But despite their mobility, most commercial quadruped robots lack manipulation capabilities. They often serve solely as mobile platforms for monitoring or exploration, equipped only with perception systems for navigation and environmental awareness \cite{fan_shi_090b9188},\cite{kaijun_wang_caa7b939}. These robots generally do not include actuators like robotic arms or grippers for interacting physically with objects.

Integrating manipulators into such systems introduces control challenges, especially when operated remotely. The operator must manage both locomotion and precise arm movements to achieve effective manipulation. This complexity is further amplified by viewpoint discrepancies between the operator and the robot's surroundings, often resulting in misjudgments of object positions and orientations \cite{florent_p__audonnet_befa8c3c}, \cite{bipasha_sen_e6c6c183}. Limited visual feedback and communication latency can also reduce control effectiveness, making object manipulation more difficult. Consequently, integrating actuators into quadruped robots necessitates more advanced and automated control systems to ensure intuitive, efficient, and reliable interaction with the environment.

Several related works have been explored in this domain. Zhang et al. \cite{QifanZhang} proposed a task-level human–robot interaction (HRI) system to support autonomous multi-objective grasping with quadruped robots. Their approach addressed the limitations in autonomous decision-making for grasping tasks by designing a touchscreen-based control terminal. This interface allowed operators to intuitively define the object search area using a video feed from the robot.

Another relevant study by Wanyan et al. \cite{LiWanyan} introduced a scene prediction and grasp pose estimation method using a YOLO-GraspNet architecture. Their method combined the fast object detection capabilities of YOLOv5s with the precise grasp pose estimation of GraspNet. 

Our previous research has focused on developing service robots to assist the elderly, including capabilities for locating lost items \cite{9172030_Billy}, autonomous navigation for wheeled platforms \cite{8973360_Zanuar}, human following \cite{8710819_Ardi}, and fall detection \cite{8124255_Agung}. To enhance human-robot interaction, we also developed pose-based activity recognition for the elderly \cite{10667962_Amik},\cite{11137561_Darmawan} . The autonomous grasping research presented in this paper complements these prior works, aiming to provide a more holistic and complete service capability.

\begin{figure*}[t]
    \centering
        \includegraphics[scale=.65, trim=0 15cm 0 0]{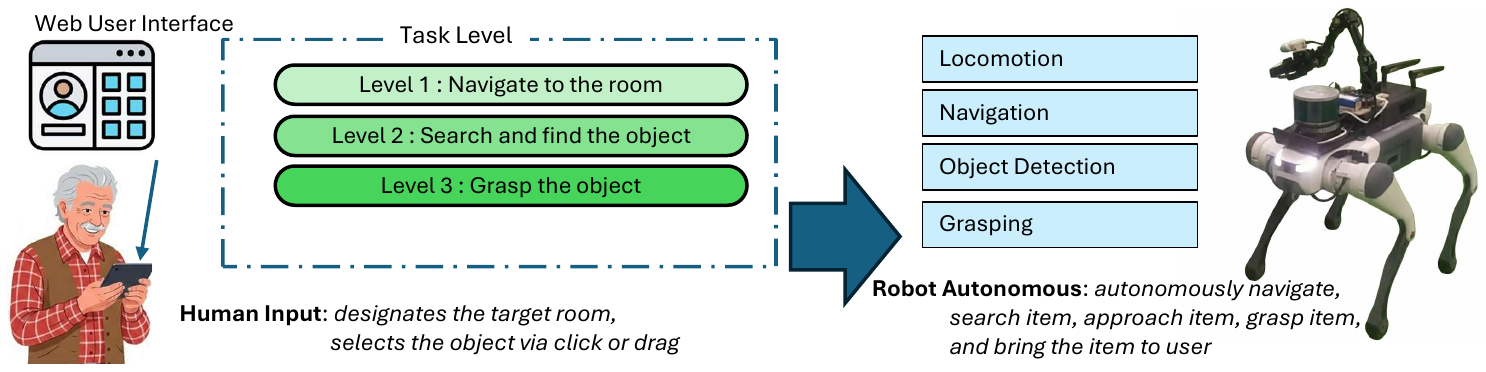}
    \caption{Overview of the proposed Task-Level Interaction system. The operator provides high-level inputs (room designation and visual object selection), which are decomposed into structured task levels. The quadruped robot then autonomously executes the corresponding low-level actions—locomotion, navigation, and manipulation—to retrieve the target object.}
    \label{fig:whole_system}
\end{figure*}

The objectives of this research are to develop a hardware integration framework that allows the attachment of a robotic arm and gripper onto a quadruped robot. This research also implement a task-level human–robot interaction system, enabling the robot to approach, detect, select, and autonomously grasp objects in a structured sequence, as described in Fig. \ref{fig:whole_system}. Then, integration state-of-the-art grasping algorithms such as GraspNet, allowing the robot arm and gripper to operate autonomously without requiring continuous manual control.
The main contributions of this study are summarized as follows:
\begin{enumerate}
    \item A modular architecture integrating the Lite3 quadruped and OpenManipulator-X, coordinated by a central embedded unit for seamless operation.

    \item An autonomous manipulation pipeline utilizing YOLOv8n and GraspNet, enhanced by a three-stage filtering strategy to derive optimal and kinematically feasible grasps .

    \item Real-world validation demonstrating effective autonomous navigation and achieving a 75\% grasping success rate.
\end{enumerate}

The remainder of this paper is organized as follows: Section II covers system design, followed by implementation details in Section III. Section IV presents experimental results, and Section V concludes the study.
\section{System Design}
\label{sec:systemdesign}

\subsection{System Logic Design}

The task-level control is designed to be accessible for operators without requiring technical programming knowledge, allowing them to select rooms, target objects, and destination points through an interactive interface, which the system translates into executable commands. This approach simplifies the control process while enhancing flexibility in adapting to real-world scenarios, as operators can easily switch mission profiles without manual code modification. Consequently, the system establishes a semi-autonomous robotic framework that fosters effective human-robot collaboration, where humans provide strategic decisions and the robot manages operational execution.

\subsection{Hardware Design}

\begin{figure} [tb]
  \centering
  \includegraphics[width=0.45\textwidth]{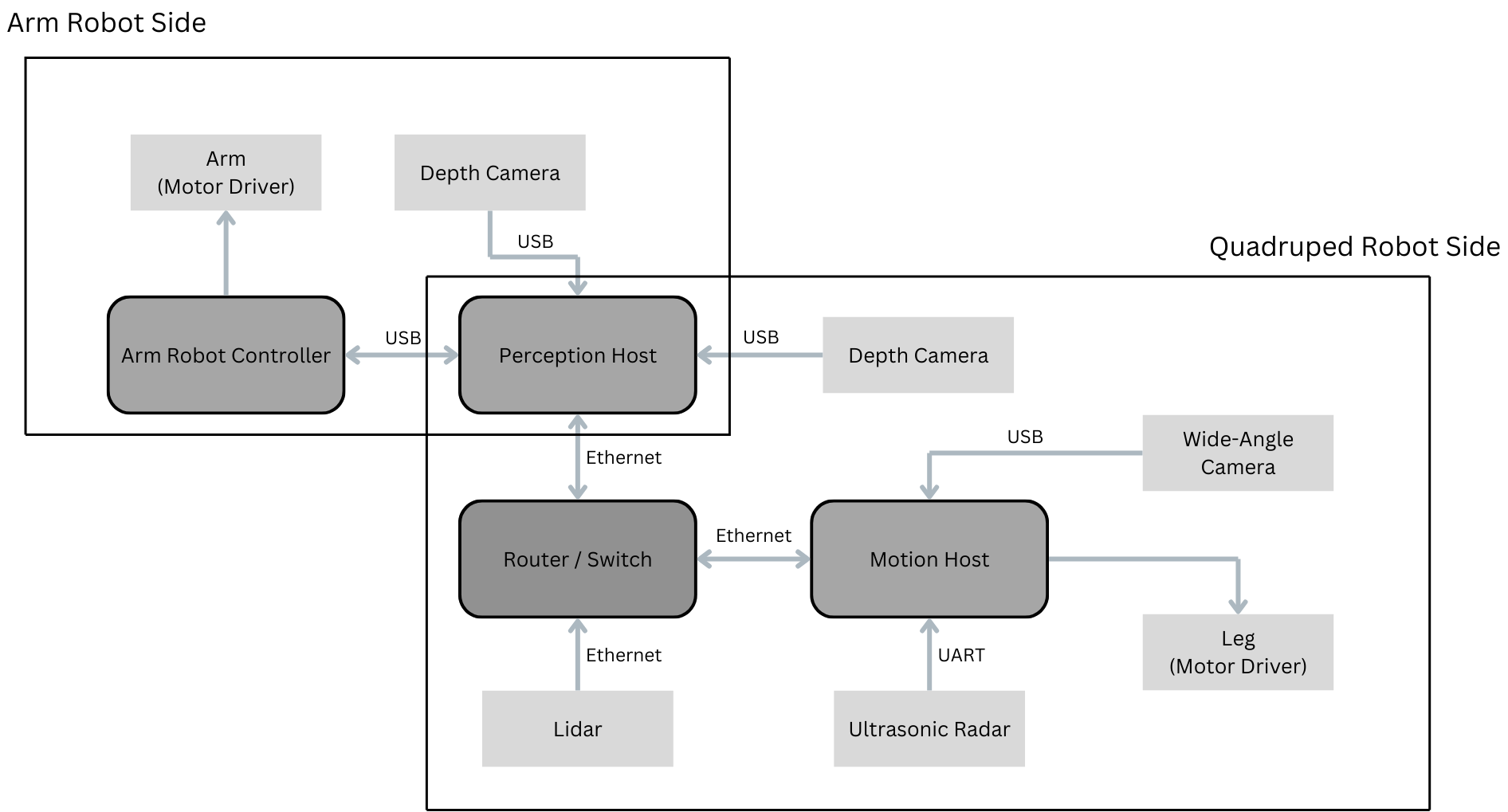}
  \caption{Hardware System Diagram}
  \label{fig:desain_sistem}
\end{figure}

In this study, a collaborative robotic system is designed to integrate two distinct types of robots—a robotic arm and a quadruped robot—working synergistically to perform manipulation and mobility tasks in a semi-autonomous manner. As illustrated in Fig. \ref{fig:desain_sistem}, the system architecture is divided into two main subsystems: the Arm Robot Side and the Quadruped Robot Side. Both subsystems are centrally coordinated by a processing unit referred to as the Perception Host. This unit utilizes the NVIDIA Jetson Orin NX\cite{NVIDIA_Jetson_Orin_ref}, a high-performance embedded computing module specifically designed for edge AI applications such as computer vision, deep learning inference, and real-time robotic control.

\subsubsection{Arm Robot Side}

On the robotic arm side, the Open Manipulator-X\cite{openmanipulator_spec_ref} is employed as the main actuator for object manipulation tasks. This lightweight, modular arm is designed for research applications and features multiple degrees of freedom driven by Dynamixel smart actuators, which provide high precision, integrated position control, and serial communication capabilities. The system is controlled by the OpenCR board\cite{OpenCR_ref}.



In addition to the actuator, the Arm Robot Side is equipped with an Intel RealSense D435i depth camera\cite{Realsense_d435_ref} to provide the system with 3D visual perception. This camera is essential for object recognition and grasping tasks. The RealSense D435i generates depth maps and point cloud data, enabling the system to estimate the position and orientation of objects accurately. The camera connects directly to the Perception Host via USB, and its output is processed locally using computer vision algorithms and AI-based object detection models. The mounting design was created using Onshape CAD software. The final design was subsequently fabricated using a 3D printer.


\subsubsection{Quadruped Robot Side}

On the quadruped robot side, the Lite3 developed by DeepRobotics is employed as a research-oriented mobility platform with robust terrain navigation and high locomotion stability. The system is controlled by a Motion Host powered by an ARM-based RK3588 processor optimized for edge computing, which manages leg motor actuation and sensor communication. The Motion Host interfaces with motor drivers and integrates data from multiple sensors, including a wide-angle camera for environmental exploration, an Intel RealSense D435i depth camera for depth-based navigation, an ultrasonic radar via UART for proximity sensing, and a LiDAR sensor via Ethernet for 360° mapping and obstacle avoidance.

\subsection{Interface Design}

In this system, the user interface is implemented using a ROS WebSocket bridge, enabling a web-based dashboard for robot control and monitoring. The interface features two live video streams: the front camera on the quadruped (left) and the gripper camera on the robotic arm’s end-effector (right). A toggle switch above the streams activates YOLOv8-based object detection, overlaying bounding boxes on detected objects in real-time. Below the video feeds, a status panel displays the robot’s current condition, such as scanning, tracking, or grasping actions.

\begin{figure} [t] \centering
  \includegraphics[scale=0.59]{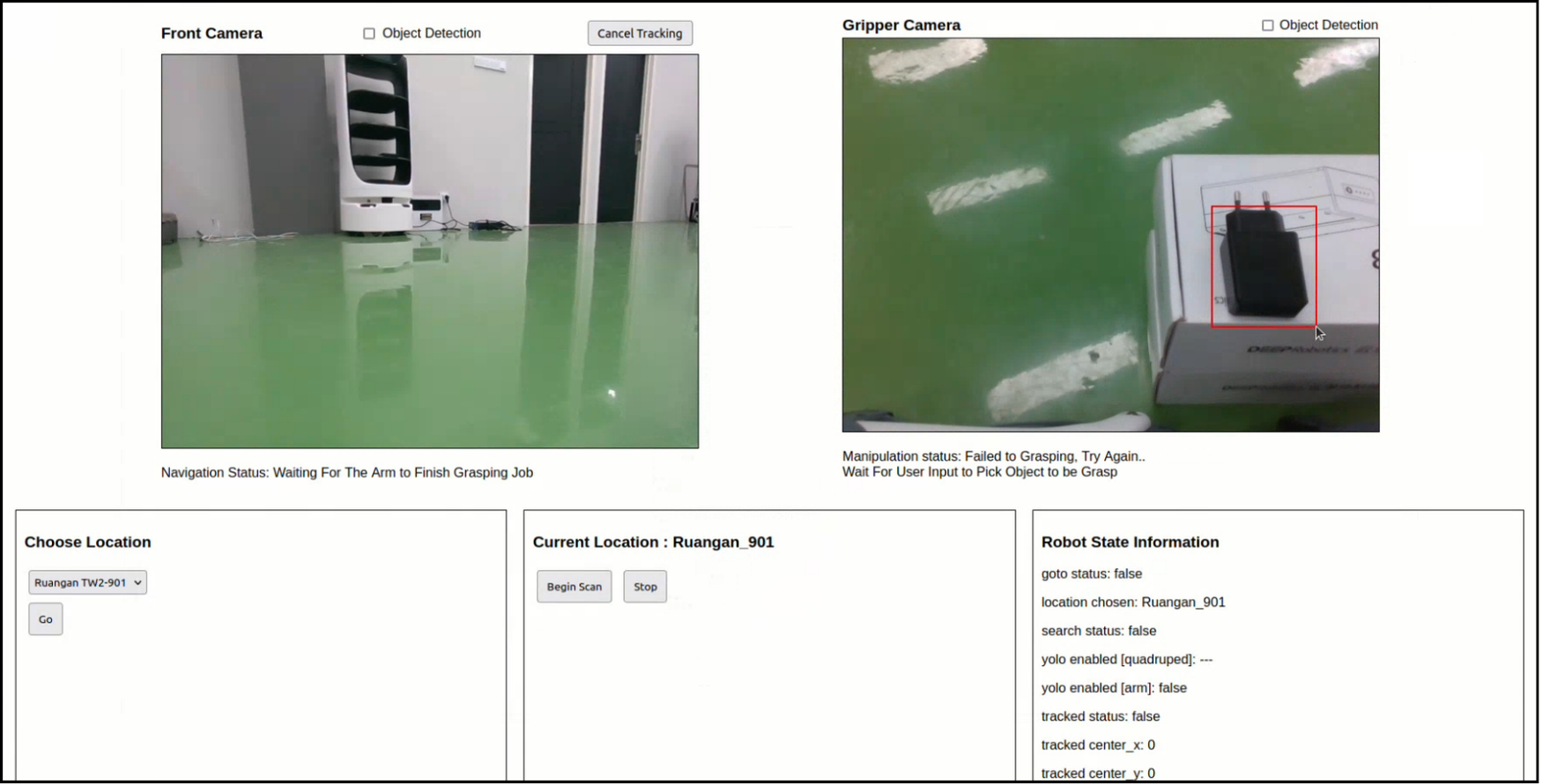}
  \caption{The web interface enables users to select target rooms (bottom-left), monitor location and control search actions (center), and view robot status (left). Live feeds from the main and arm-mounted cameras are displayed at the top.}
  \label{fig:web_interface_edited}
\end{figure}

The lower section is organized into three functional blocks. The left block allows target room selection via a dropdown menu and a "Go" button to initiate navigation. The middle block shows the robot’s current location and includes "Begin Scan" and "Stop" buttons to start or stop circular searches. The right block provides real-time operational status, indicating whether the robot is searching, has detected the target object, or is performing specific tasks. This layout ensures intuitive control and monitoring aligned with the defined task-level architecture.
  \section{Implementation}
\label{sec:implementation}

\begin{figure} [t] \centering
  \includegraphics[scale=0.15]{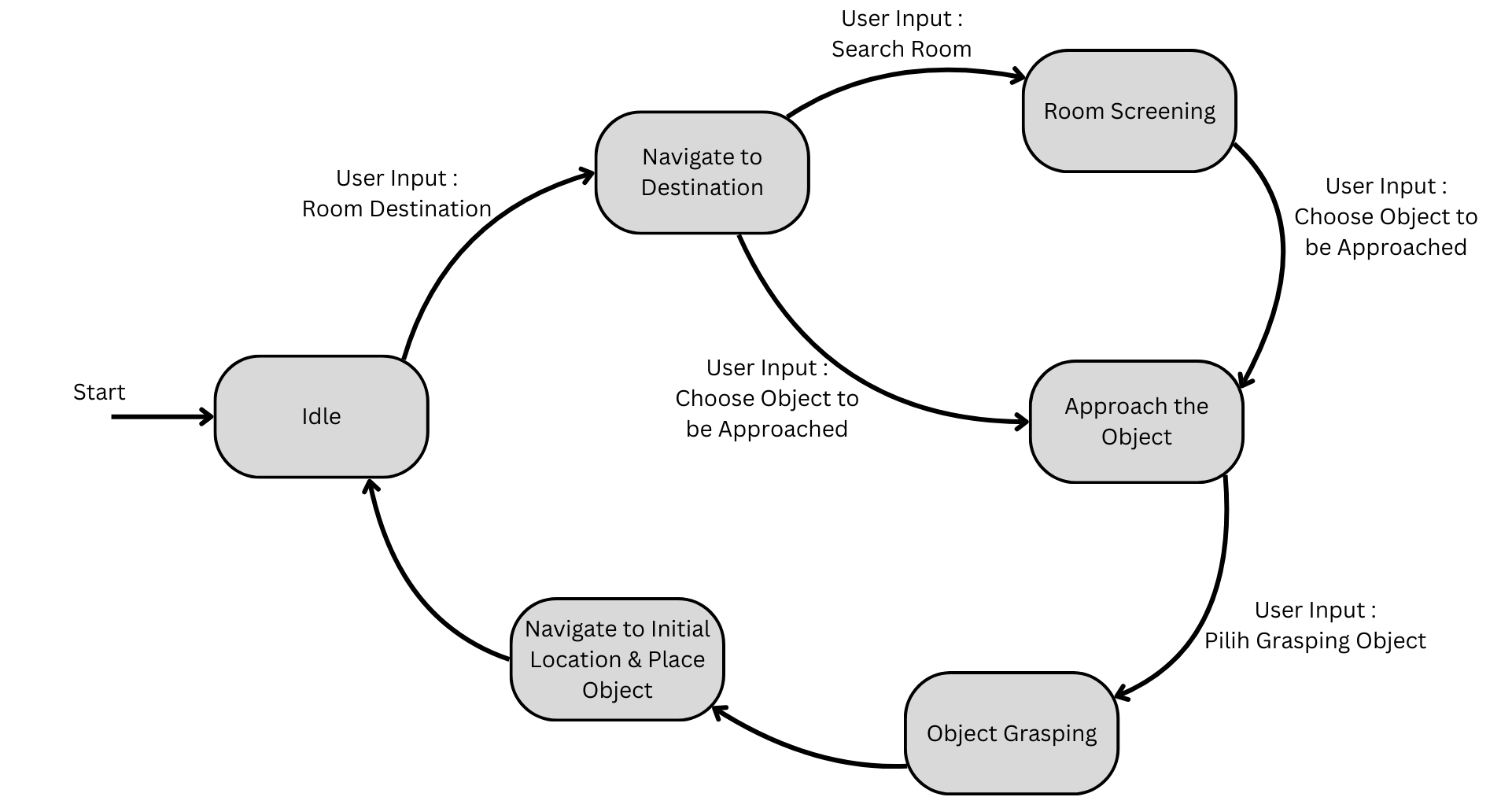}
  \caption{FSM Flow System}
  \label{fig:fsm_flow_robot}
\end{figure}

The robotic system is designed to execute a complete task from start to finish by decomposing the task into multiple structured stages, organized within a finite state machine (FSM), as illustrated in Fig. \ref{fig:fsm_flow_robot}. This FSM-based approach enables the system to modularly coordinate robotic actions related to both navigation and manipulation, while maintaining structured transitions across various phases. Moreover, it provides flexibility for user interaction at each state, reinforcing the robot's semi-autonomous capabilities.

The FSM structure enables the robotic system to follow a systematic and sequential workflow while maintaining user control at key stages. This architecture not only simplifies system coordination but also provides adaptability for more complex and dynamic real-world scenarios. To summarize, the robot executes the following sequential steps:

\begin{enumerate}
    \item The robot autonomously navigates from the initial point to the user-selected destination room.
    \item The robot scans the room to detect objects. Once objects are identified, the user selects one, prompting the robot to approach and sit beside the object.
    \item Using the camera mounted on the robotic arm, the user selects the specific object to be grasped, triggering the arm to execute a grasping operation.
    \item The robot transports the grasped object back to the initial point and places it at the designated location.
\end{enumerate}

\subsection{Navigation System}

\begin{figure} [t] \centering
  \includegraphics[scale=0.17]{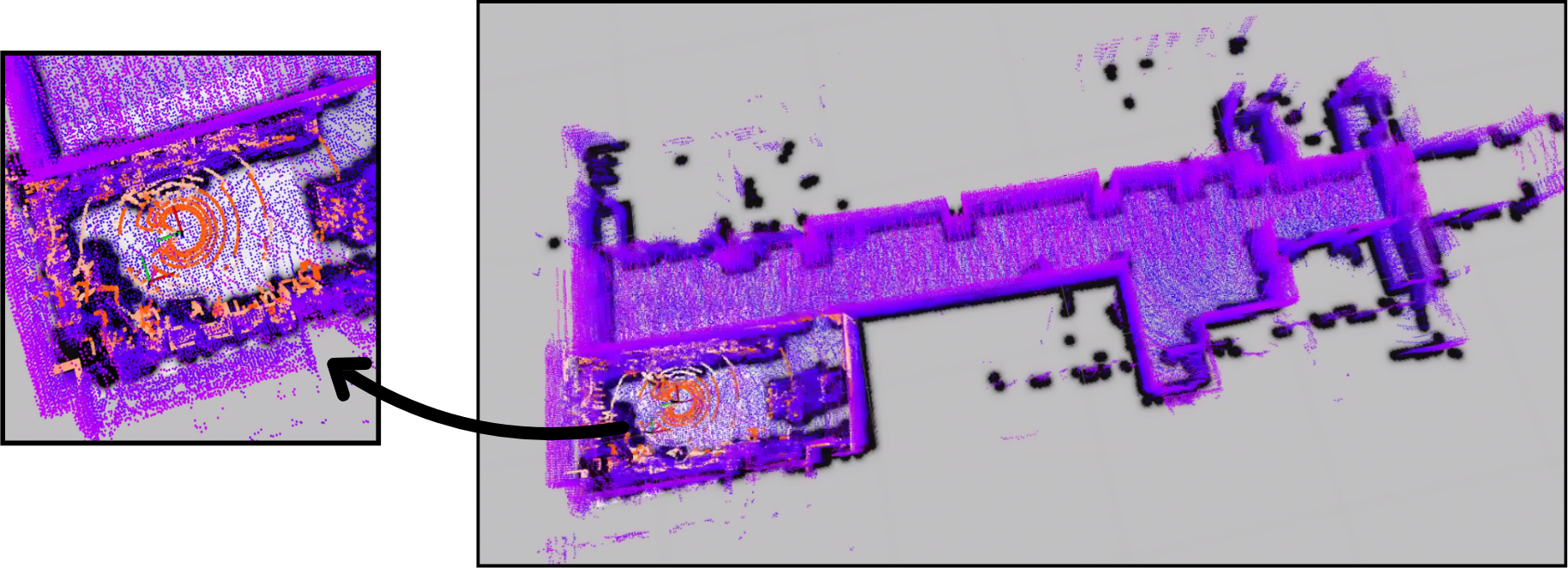}
  \caption{Mapping Result}
  \label{fig:mapping_result}
\end{figure}

In this study, mapping and localization utilize hdl\_graph\_slam and hdl\_localization, proposed by Koide et al. \cite{koide_slam}, using LiDAR point clouds. Fig. \ref{fig:mapping_result} illustrates the resulting map of the 9th floor of Tower 2 at Institut Teknologi Sepuluh Nopember (ITS).

\begin{figure} [t] \centering
  \includegraphics[scale=0.4]{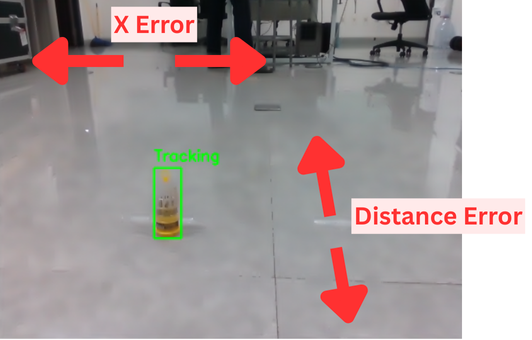}
  \caption{Object chosen and robot attempts to approach it.}
  \label{fig:nav_deteksi_objek2}
\end{figure}

In this study, the navigation system integrates object detection and tracking as the basis for movement decisions. Object detection is performed using the YOLOv8n model \cite{yolov8_ref}, enabling real-time recognition of multiple objects from RGB camera frames. Users can select the target object either by clicking on it or using drag-select to handle detection inconsistencies, after which the system switches from detection to tracking mode using cv2.TrackerCSRT\_create(), which is robust to scale variation and partial occlusion. Real-time path planning and obstacle avoidance are managed by the ROS Navigation Stack, while a PID controller adjusts linear and angular velocities based on visual feedback (\textit{x} and distance errors) from the tracker.

The outcome of this navigation process is that the robot is positioned in close proximity to the selected object and is in a seated posture. This position is considered optimal for executing manipulation tasks, such as object grasping or placement, the object used for demonstration is a yellow bolt set.

\subsection{Manipulation System}

After the quadruped robot completes the navigation task and positions itself in a seated posture near the target object, the manipulation process is initiated.

\begin{figure} [t] \centering
  \includegraphics[scale=0.25]{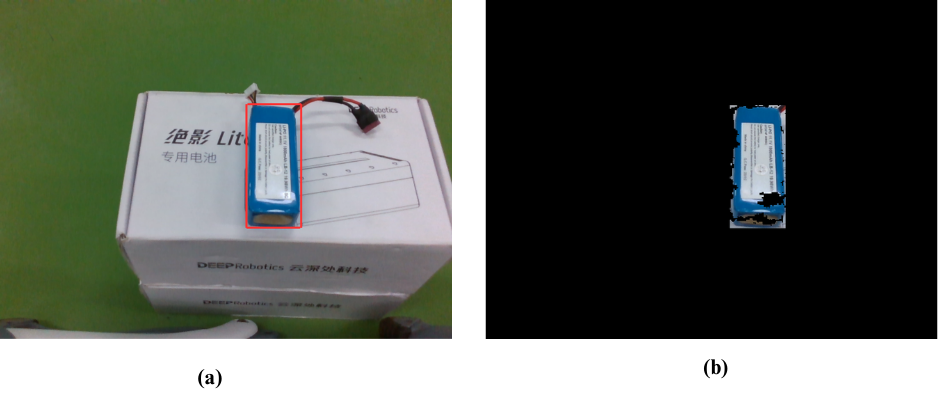}
  \caption{(a) Selected object, (b) masking result}
  \label{fig:pilih_objek}
\end{figure}

The manipulation sequence starts with the robotic arm moving to its highest reachable position, enabling a wider camera field of view. The onboard camera then streams RGB video, upon which the YOLOv8n model is applied to detect surrounding objects. As shown in Fig. \ref{fig:pilih_objek}.a, detected objects are enclosed in bounding boxes, and users may select a specific object for manipulation by directly clicking on it within the interface. Alternatively, a drag-select feature is provided to address cases where YOLO detection results are inconsistent. In this mode, users can draw a rectangular region around the desired object, prompting the system to confirm the selection through a pop-up notification. This confirmation step allows users to retry the selection until the intended object is accurately highlighted.

Upon confirming the selected object, the system captures and stores three essential input files: the RGB color frame, the aligned depth frame, and the intrinsic camera matrix. These inputs are prerequisites for generating grasp candidates using the GraspNet framework. Notably, GraspNet operates agnostically to object identity and location—it generates grasp poses across the entire input scene. To ensure that the grasp poses are generated only around the selected object, a masking process is applied. Specifically, only the pixels within the selected bounding box are retained, as shown in Fig. \ref{fig:pilih_objek}.b, ensuring that GraspNet focuses exclusively on the target object.

\begin{figure} [t] \centering
  \includegraphics[scale=0.2]{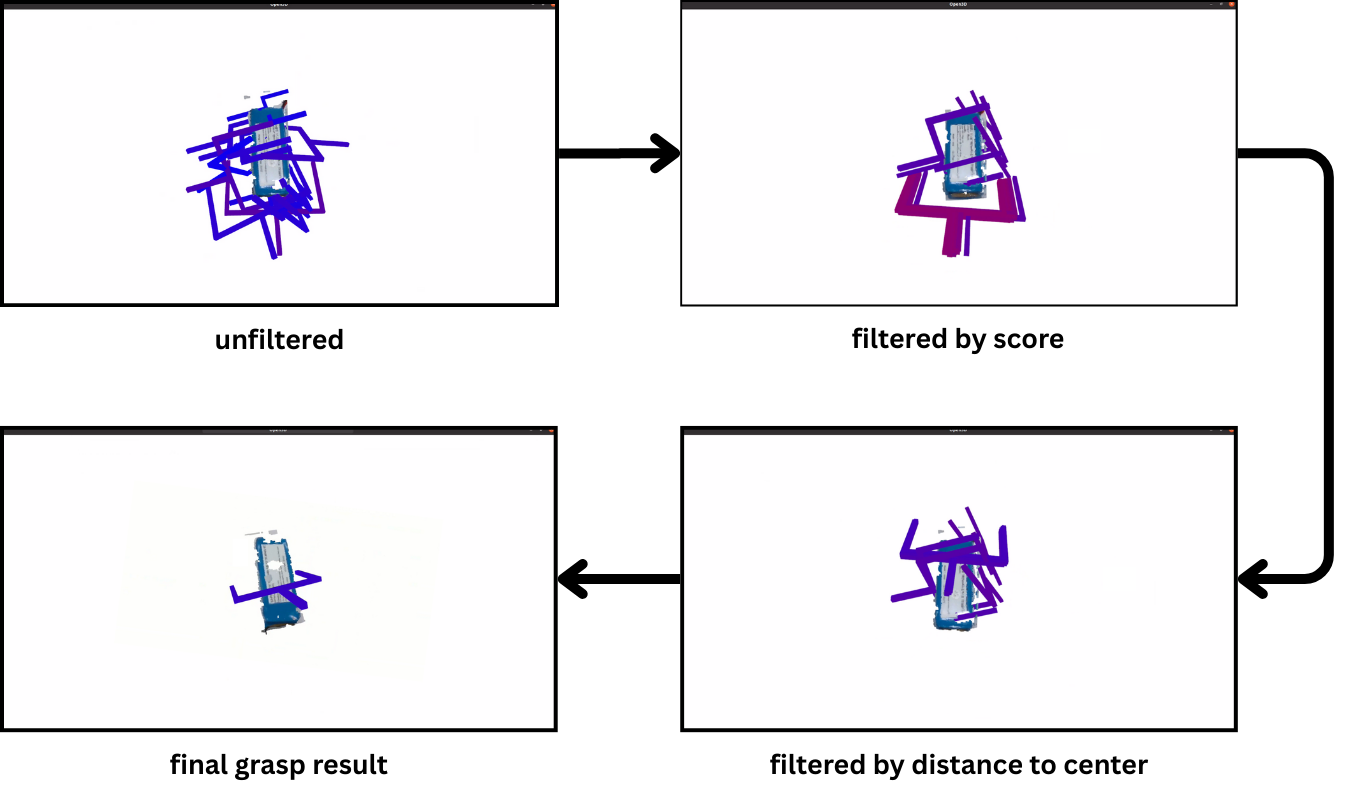}
  \caption{Grasp poses filtering process}
  \label{fig:grasp_filtering_img}
\end{figure}

Subsequently, the color frame, depth frame, and camera matrix are passed into GraspNet. The framework outputs a set of grasp candidates, each consisting of pose parameters and grasp quality metrics. A visual representation of the resulting grasp poses is provided using the Open3D library in Fig. \ref{fig:grasp_filtering_img}. From the numerous grasp poses generated by GraspNet, only a single optimal grasp is selected for execution. This necessitates a filtering process to identify the most feasible and reliable grasp configuration for the robotic arm.

The first criterion is the confidence score, which reflects the likelihood of successful grasp execution based on the geometric structure of the object and the prediction output of the GraspNet deep learning model. A higher score indicates a higher probability that the gripper can stably grasp the object without slipping, missing, or destabilizing it. In this stage, the top 20 grasp poses with the highest confidence scores are selected, effectively eliminating low-confidence candidates that may lead to failure.


The second criterion is the distance between each grasp pose and the object’s center point. In precision grasping, poses that deviate significantly from the object's center of mass are likely to result in unbalanced or unstable grasps. Among the remaining candidates, the pose closest to the object’s center point is selected, as it ensures a more physically secure and symmetrical grasp location on the target object.


The final step involves orientation adjustment of the selected grasp pose to align it with the actual physical constraints and capabilities of the robotic arm. Although GraspNet outputs a full 6-DOF pose, the orientation might not be immediately feasible due to the kinematic limits or joint configurations of the manipulator. Therefore, a rotation is applied to the grasp pose matrix to transform it into a reachable and executable orientation that avoids extreme joint angles or motion discontinuities.


Through this three-stage filtering process, the system identifies a single, high-confidence grasp pose that is both physically feasible and optimal in terms of grasp stability. This final grasp pose is then used as the primary input for the subsequent motion planning phase, enabling the robot to execute manipulation tasks with a high probability of success.


The grasping pose estimated by GraspNet is defined relative to the camera frame, not the robot’s frame. This means the position and orientation—comprising translation and rotation values—are provided in the coordinate system of the camera, with the origin at the camera lens and axes aligned to the camera’s orientation. While this representation is useful for interpreting the object’s spatial properties from the camera’s perspective, it is insufficient for direct use in robotic motion execution. To enable robotic manipulation, a transformation is required to convert the grasping pose from the camera frame to the robot’s base or world frame. This transformation incorporates both translation and rotation to accurately represent the camera's position and orientation with respect to the robot. The grasping pose from GraspNet (camera frame) is transformed to the robot base frame using the manipulator's kinematic model described in \cite{ZichangZhou}.


\begin{figure} [t] \centering
  \includegraphics[scale=0.04]{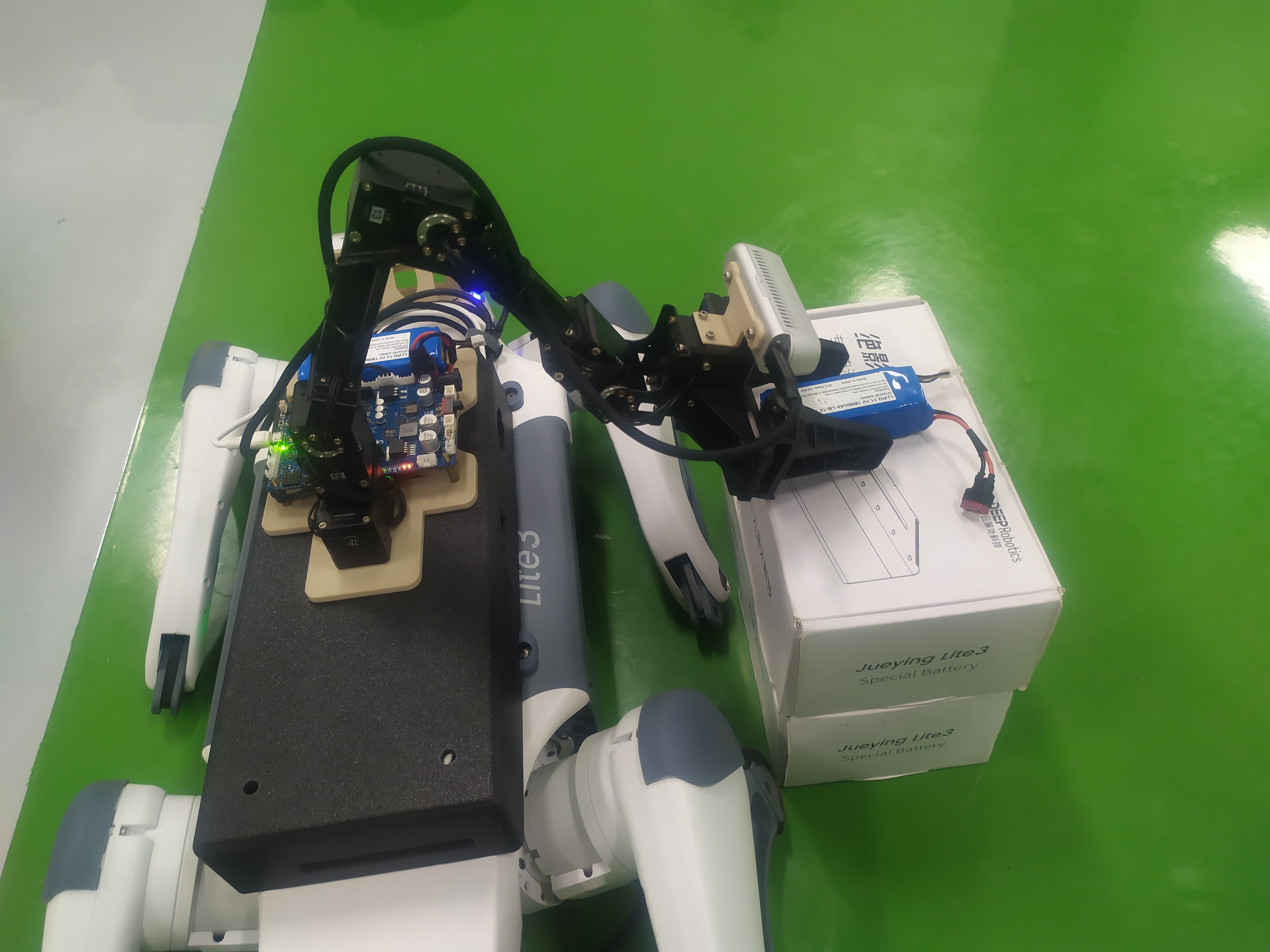}
  \caption{Robot performing the grasping action.}
  \label{fig:foto_grasping}
\end{figure}

Once the grasping pose is transformed into the robot’s frame using the robot’s kinematic configuration, the pose data is passed to the motion planning module, which generates a trajectory for the manipulator to reach and execute the grasping task efficiently and precisely.

\section{Experiments}
\label{sec:experiments}

To validate the proposed system and evaluate its performance in real-world scenarios, a series of experiments were conducted involving both navigation and object manipulation tasks. The experiments are designed to demonstrate how the integrated system—comprising object detection, autonomous navigation, visual tracking, grasp planning, and motion execution—functions cohesively in completing tasks from user instruction to object delivery. Each phase of the robot's behavior is tested under various conditions to examine its robustness, accuracy, and responsiveness in semi-autonomous operations. The results provide insights into the strengths and limitations of the proposed approach, as well as potential directions for future improvements.

The evaluation of the robot's movement in approaching objects was conducted to assess the accuracy of the navigation and control system in tracking and approaching the object selected by the user. In terms of final position accuracy, it was observed that the position of the object relative to the robot varied but remained within a relatively close range, allowing for successful manipulation. The measured final position values indicated that the object was generally located in front of and slightly to the right of the robot, with coordinates ranging from (24 cm, 5 cm) to (27 cm, 16 cm). The distance from the robot to the object, calculated from the final (\textit{x}, \textit{y}) positions, ranged between 24.52 cm and 33.42 cm.

\begin{table}[]
\caption{Experiment on Object Grasping}
  \label{tb:experiment_on_object_grasping}
\begin{xtabular}{|c|c|c|c|c|}
  
  \hline
  \rowcolor[HTML]{C0C0C0}
  \textbf{n} & \textbf{Object} & \textbf{Method} & \textbf{Duration} & \textbf{Grasp Status} \\
  \hline
  1 & Charger &  drag & 64 seconds & success \\
  2 & Charger &  drag & 61 seconds & success \\
  3 & Charger &  click & 60 seconds & success \\
  4 & Charger &  click & 58 seconds & success \\
  5 & Golf Ball &  click & 20 seconds & fail \\
  6 & Golf Ball &  click & 55 seconds & success \\
  7 & Golf Ball &  click & 59 seconds & success \\
  8 & Golf Ball &  click & 17 seconds & fail \\
  9 & Battery &  click & 63 seconds & success \\
  10 & Battery & click & 57 seconds & success \\
  11 & Battery & click & 24 seconds & fail \\
  12 & Battery & click & 61 seconds & success \\
  \hline
\end{xtabular}
\end{table}


The testing in this section was conducted to evaluate the robotic manipulation system’s capability in autonomously performing object grasping and placement tasks using the robot arm and gripper. The focus of the evaluation was to measure the effectiveness and accuracy of the system throughout the grasping process, starting from object detection to the final placement.

The grasping tests yielded promising results, with the robot achieving a success rate of 75\% across 12 trials. High success rates were observed with stable objects such as a charger. However, the system encountered significant challenges with small, slippery objects like a golf ball and heavier items such as a battery. These findings indicate the need for further improvement in grasping force and gripper reliability during manipulation tasks.

In the failed trials, specific challenges were identified based on the object characteristics and grasping conditions. In trial 5, the grasp attempt on the golf ball failed because the ball slipped during grasping, as its spherical shape provided less surface stability compared to the box-shaped charger, making it prone to slipping. Similarly, in trial 8, the grasping failed when the golf ball rolled away after being slightly nudged by the gripper. This again highlights the inherent difficulty of handling spherical objects, which do not remain stationary when disturbed. Lastly, in trial 11, the grasping failure occurred with the battery, which is heavier than the other tested objects. The improper grasping position—closer to the edge rather than the center—combined with the object’s weight, resulted in the gripper’s inability to securely hold it.

\section{Conclusion and Future Work}
\label{sec:conclusion}

Based on the experimental results obtained in this study, several key conclusions can be drawn. The quadruped robot, when integrated with a robotic arm and vision system, successfully performed a series of service-level tasks in a task-level approach. These tasks included autonomous navigation to a selected location, object detection and grasping, and accurate placement of the object. The system demonstrated consistent and reliable performance throughout the execution of these tasks. Coordination between the quadruped robot and the robotic arm was effectively established through the addition of a mini PC, which acted as a central processing unit to manage communication between both subsystems. The implementation of the GraspNet algorithm for object manipulation showed encouraging results, achieving a grasp success rate of 75\% across 12 trials. High success rates were recorded for stable objects such as a charger, while performance decreased when handling smaller, more slippery objects like golf balls or heavier items such as batteries. These challenges suggest the need for future improvements in grasp strength and control precision. Overall, the system developed in this study demonstrates a viable approach to semi-autonomous task-level control in quadruped mobile manipulation.

While the robot has successfully demonstrated the capability to search, detect, grasp, and deliver objects, future work will focus on conducting comprehensive user studies to evaluate the robot's overall performance and analyze user preferences regarding object selection methods. Additionally, we plan to integrate Large Language Models (LLMs) into the system to enable more intuitive control through natural language commands.


\section*{Acknowledgment}

This research was financially supported by the Final Project Assistance Grant (Bantuan Tugas Akhir Mahasiswa) funded by Institut Teknologi Sepuluh Nopember (ITS) under the 2025 ITS Internal Research Grant Scheme.

  \bibliographystyle{IEEEtranN}
  \bibliography{pustaka/pustaka.bib}

  \balance

\end{document}